\documentclass{article}
\usepackage[T1]{fontenc}
\usepackage{iclr2027_conference,times}
\usepackage{amsmath,amssymb,booktabs,graphicx,microtype}
\usepackage{array,colortbl}
\usepackage{flafter}
\usepackage{placeins}
\usepackage{longtable}
\usepackage{float}
\usepackage{algorithm}
\usepackage[noend]{algpseudocode}
\usepackage{hyperref}
\usepackage{url}
\definecolor{planblue}{HTML}{174A78}

\hypersetup{pdftitle={Plan Canvas: Fixed Reasoning Regions for Continuous Language Flows}}

\title{Plan Canvas: Fixed Reasoning Regions\\for Continuous Language Flows}

\author{%
\parbox{\dimexpr\textwidth-2\tabcolsep\relax}{%
\centering
{\normalsize\bfseries
Miaohe Niu$^{1}$, Pengxiang Li$^{2}$,
Jingbo Zhu$^{1,3}$, Tong Xiao$^{1,3,*}$}\\[4pt]
{\normalfont\small
$^{1}$School of Computer Science and Engineering,\\
Northeastern University, China\\
$^{2}$Hong Kong Polytechnic University, China\\
$^{3}$NiuTrans Research, Shenyang, China\\[3pt]
\texttt{niumiaohe@mails.neu.edu.cn},
\texttt{pengxiang.li@connect.polyu.hk}\\
\texttt{xiaotong@mail.neu.edu.cn}\\[2pt]
$^{*}$Corresponding authors.
}%
}%
}

\iclrfinalcopy

\begin{document}
\maketitle
\pagestyle{plain}
\thispagestyle{plain}

\begin{abstract}
Continuous language flows generate text by denoising all positions of a target canvas together.
The natural way to add reasoning to such a model is to write a trace ahead of the answer, but the trace length changes from question to question.
The answer start is therefore unknown during denoising, and the model has to decide the trace length, the place of every trace token, and the answer at the same time.
We propose Plan Canvas to fix the boundary between the trace and the answer.
A plan region of fixed capacity holds a compact trace, supervised padding fills its unused positions, and the answer starts at a fixed position.
The fixed regions also allow separate denoising clocks for the plan and for the answer.
With the trace text, backbone, and canvas length of the free-trace baseline held fixed, Plan Canvas improves accuracy on ProsQA and on Deep ProsQA, a graph benchmark with longer proofs.
On Deep ProsQA, accuracy rises from 73.0\% to 87.0\%, the share of questions answered with a valid path rises from 30.8\% to 59.1\%, and the gain is largest on the longest proofs.
\end{abstract}

\section{Introduction}

Language models answer multi-step questions more accurately when they write intermediate steps before the answer~\citep{wei2022chain,wang2023selfconsistency,zhou2022least,wang2023plan,wang2024chain}.
In autoregressive decoding, this layout costs little.
The model writes the trace one token at a time and decides where the trace ends, and the answer begins at the next position.
Continuous language flows generate text in a different way~\citep{hu2026elf}.
A continuous language flow starts from noise on a canvas of embedding vectors and updates every target position at every denoising step.
There is no left-to-right order, so the layout of the output has to be settled in another way.

The direct way to add reasoning to a continuous language flow is to copy the autoregressive layout and write the trace into the target ahead of the answer.
We call this layout a free trace.
In a free trace, the answer start depends on the trace length, and the trace length is part of what the model is still generating.
The model has to decide the length of the trace, the place of every trace token, and the answer together, and a change in any one of them moves the others.
On graph reasoning, a free trace improves accuracy over direct answering, but many emitted traces do not form a valid path in the graph.
We argue that part of this gap comes from the moving answer boundary and not from the trace text.

Earlier parallel generators fix parts of the output layout.
Semi-autoregressive and block diffusion models generate the output block by block~\citep{han2023ssd,arriola2025blockdiffusion}.
Planned Diffusion first predicts the response chunks and then denoises the chunks in parallel~\citep{israel2025planned}, and ReFusion alternates slot selection with autoregressive infilling~\citep{li2025refusion}.
\citet{yeom2026answer} study when a diffusion language model commits to its answer relative to its reasoning text.
These methods control where generated text goes or when the answer is committed, but none of them fixes the boundary between a supervised reasoning trace and its answer inside one continuous flow.

Plan Canvas fixes that boundary.
Plan Canvas reserves a plan region of fixed capacity for a compact trace, fills the unused plan positions with supervised padding, and starts the answer at a fixed position.
The trace is still supervised text, the trace still comes before the answer, and the model still denoises both regions jointly with bidirectional attention.
Only the layout changes, so the trace length no longer decides where the answer starts.
The fixed regions also allow a second change.
Every position belongs to a known region, so the plan region and the answer region can each receive their own denoising time.
We call these times segment clocks.
With segment clocks, one trained model can denoise the plan and the answer together or let one region lead, and the order of plan and answer becomes a choice at inference.

We compare Plan Canvas with direct answering, the full trace, the free trace, and an answer-then-trace control on ProsQA and on Deep ProsQA, a graph benchmark with longer proofs that we build for this study.
All layouts share the trace text, the backbone, and the canvas length, so the comparison changes only where the trace and the answer sit on the canvas.
Plan Canvas is more accurate than the free trace on both benchmarks, the paths it emits pass a graph check far more often, and its advantage is largest on the longest proofs.
Free traces often repeat parts of the path they have already written, and the fixed plan region removes most of these repeats.

The paper makes three contributions.
First, we identify the moving answer boundary of a free trace as a cost of writing reasoning into a jointly denoised canvas, and we separate this cost from the trace text with matched layout controls in Section~\ref{sec:experiments}.
Second, we introduce Plan Canvas and segment clocks in Section~\ref{sec:method}.
Third, we show on graph reasoning that the fixed boundary improves both answer accuracy and the validity of the emitted paths, and we measure how the order of plan and answer affects the result.

\section{Plan Canvas}
\label{sec:method}

\subsection{The layout problem in a jointly denoised target}

A continuous language flow denoises the whole target at once~\citep{hu2026elf}.
For a prompt $q$ with $n$ tokens, the target occupies the remaining positions of a canvas with $S$ positions, and every target position holds an embedding vector that moves from noise to a clean state.
A free trace writes the trace tokens $p_1,\ldots,p_m$ directly after the prompt and the answer $a$ directly after the trace.
The answer then begins at position $n+m$, and $m$ differs across examples.
During denoising the model has to resolve $m$, place each trace token relative to an end that is not yet resolved, and form the answer at a position that moves with $m$.
Plan Canvas removes $m$ from the layout: the trace receives a region of fixed capacity, and the answer receives a fixed start.
Fixed regions also tell the model the role of every position before generation starts.
Section~\ref{sec:clocks} uses this to give each region its own denoising clock.

\subsection{A fixed plan region for a compact trace}

The plan region holds a compact trace: the intermediate computations that connect the question with the supervised answer.
On graph reachability, the trace lists the concept names visited from an entity fact to the correct candidate.
Tokenizing the trace gives the plan $p=(p_1,\ldots,p_m)$.

The trace goes into a region of fixed capacity $K$ directly after the question, and padding fills the positions the trace does not use:
\begin{equation}
 u = [q;\ p_1,\ldots,p_m;\ \mathrm{PAD}^{K-m};\ a;\ \mathrm{EOS},\ldots,\mathrm{EOS}],
 \qquad |u|=S.
 \label{eq:canvas}
\end{equation}
The capacity $K$ is shared by all examples of a dataset and covers every training trace by default.
We set $K$ by rounding the maximum training trace length up to the next multiple of eight tokens.
For a prompt with $n$ tokens, the answer always starts at position $n+K$, whatever the trace length.
The internal PAD tokens receive supervision like any other target token, so the model learns where the trace ends and leaves the rest of the region empty.
The positions after the answer receive EOS supervision.

A readout window fixes how many target positions are decoded.
We use $G=S-Q_{\max}$, where $Q_{\max}$ is the fixed prompt cap of the dataset.
Only the first $G$ target positions are read, and the first generated EOS ends the decoded text.
Sampling evolves every non-prompt position of the full canvas, including the trailing positions beyond the readout window.
The comparison with a free trace holds both $G$ and $S$ fixed within each dataset, so the two layouts differ only in where the trace and the answer are placed.

\subsection{Continuous states and segment clocks}
\label{sec:clocks}

Flow states hold one frozen-encoder vector per canvas position throughout denoising.
We use ELF with a frozen T5-small encoder and normalize its contextual embeddings~\citep{hu2026elf,raffel2020t5}.
The evolving vectors have 512 dimensions, and the input projection of the denoiser passes through a bottleneck of 128 dimensions.
Prompt embeddings depend only on prompt tokens, while training target embeddings contextualize the complete supervised target.
Prompt states stay fixed during flow training, and sampling initializes every target state with Gaussian noise.

In a fixed layout the region of a position is known before generation, so each region can carry its own denoising time.
Let $s(i)\in\{p,a\}$ give the region of target position $i$, with EOS positions assigned to the answer region.
The corrupted state at position $i$ follows the interpolation of the underlying flow model:
\begin{equation}
 z_i(t_{s(i)})=t_{s(i)}x_i+(1-t_{s(i)})\sigma\epsilon_i,
 \qquad \epsilon_i\sim\mathcal{N}(0,I),
 \label{eq:interpolation}
\end{equation}
where $x_i$ is the clean embedding and the shared noise scale is $\sigma=2$.
Time zero denotes a fully noisy state, and time one denotes the clean endpoint for either region.

The denoiser reads each region's clock through time-prefix tokens and a position-dependent time term.
After projecting the noisy state and the self-conditioning input~\citep{chen2022analog}, the model adds the segment information:
\begin{equation}
 h_i = P([z_i;\hat{x}^{\mathrm{prev}}_i]) + e_{s(i)} + W_t\phi(t_{s(i)}).
 \label{eq:conditioning}
\end{equation}
Here $e_s$ is a learned segment marker, and the affine time projection $W_t$ starts at zero.
Each region also receives a group of time-prefix tokens with the same marker and time embedding.
Attention stays bidirectional, so plan and answer states exchange information in every joint update.
The single-clock variant keeps the original ELF conditioning interface on the same fixed layout, so the comparison of the two variants separates the effect of the layout from the effect of the clocks.

\begin{figure}[htbp]
\centering
\includegraphics[width=\linewidth]{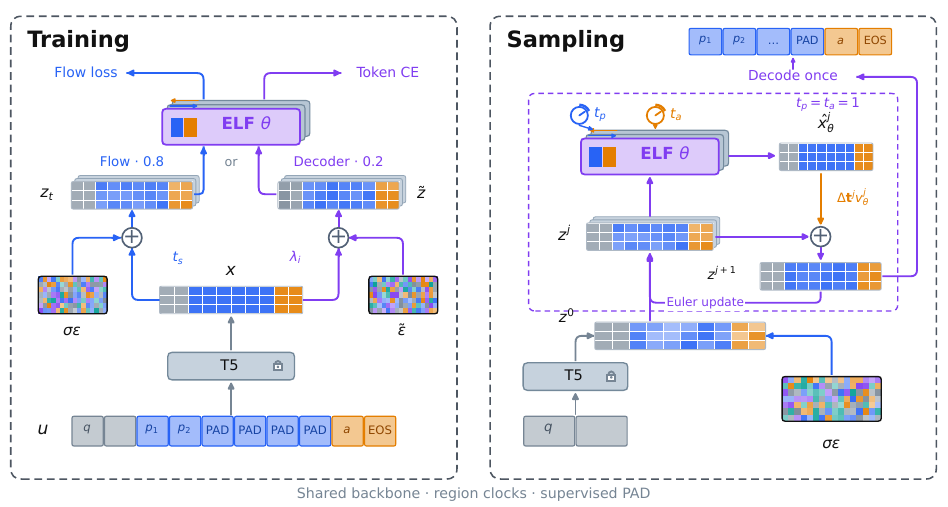}
\caption{Training and sampling use the same ELF backbone with fixed plan and answer regions.
Each training example selects either flow matching with region corruption or token cross-entropy with separate decoder corruption.
The branch probabilities are 0.8 for flow matching and 0.2 for token cross-entropy in the reported configuration.
During sampling, the prompt stays fixed while region clocks control Euler updates of every target position.
The model predicts clean states, which define velocity, and switches to token decoding once both regions reach time one.}
\label{fig:framework}
\end{figure}

\subsection{Training on the fixed layout}

Training keeps the flow and decoder objectives of ELF and extends supervision to the padding.
Figure~\ref{fig:framework} shows the shared training branches and the iterative sampling procedure.
For flow examples, clean-state prediction defines velocity through the clipped denominator used in ELF:
\begin{equation}
 v_{\theta,i}=\frac{\hat{x}_{\theta,i}-z_i}{\max(1-t_{s(i)},\tau)},
 \qquad
 v_i^{\star}=\frac{x_i-z_i}{\max(1-t_{s(i)},\tau)},\quad \tau=0.05.
 \label{eq:velocity}
\end{equation}
The implementation keeps the self-conditioning correction to the velocity target.
Appendix~\ref{app:objective} describes the correction.
Decoder examples reconstruct discrete tokens from separately corrupted embeddings with the shared Transformer and a vocabulary head.
Each example selects decoder training with probability 0.2 and flow training with probability 0.8.
Both losses exclude prompt positions and include internal PAD tokens and trailing EOS positions.

Training times decide which combinations of plan and answer noise the segment-clock model sees.
With probability $\rho$, both regions receive the same time from the shared logit-normal distribution.
With probability $1-\rho$, their times are drawn independently from that distribution.
The default segment-clock model uses tied training times with $\rho=1$.
The mixed-time variant sets $\rho=0.5$, so half of its training examples keep equal plan and answer times.

\subsection{Offset schedules at inference}

Two clocks let the regions advance at different times with the same trained weights, so the order of plan and answer becomes an inference choice.
Let $0=t_0<\cdots<t_N=1$ be the sampled and sorted time grid shared by both regions.
At global iteration $j$, region $s$ uses index $k_s(j)=\min(\max(j-o_s,0),N)$ for offset $o_s$.
Each joint velocity evaluation predicts both regions, and each active region then takes its own Euler update:
\begin{equation}
 z_i^{j+1}=z_i^j+\Delta t_{s(i)}^j\,v_{\theta,i}^j,
 \qquad
 \Delta t_s^j=
 \begin{cases}
 t_{k_s(j)+1}-t_{k_s(j)},&o_s\leq j<o_s+N,\\
 0,&\text{otherwise}.
 \end{cases}
 \label{eq:euler}
\end{equation}
An inactive region keeps its current state and stays visible to the region being updated.
After both regions reach their endpoints, the shared decoder maps the complete canvas to tokens once.
We extract the answer from the decoded text and keep the emitted trace for diagnostics.

Synchronous offsets $(0,0)$ finish in $N$ joint iterations and form the default schedule.
A plan lead of $(0,N/2)$ uses $3N/2$ iterations and leaves an interval of overlapping updates.
Plan-first offsets $(0,N)$ and answer-first offsets $(N,0)$ each use $2N$ joint iterations.
Every schedule is therefore reported with its joint iteration budget next to the active step count per region.
Algorithm~\ref{alg:sampling} summarizes the sampling procedure.
The single-clock variant is the special case with one region time and zero offsets.

\begin{algorithm}[t]
\caption{Sampling with segment clocks and offsets $(o_p,o_a)$}
\label{alg:sampling}
\begin{algorithmic}[1]
\Require prompt $q$, time grid $0=t_0<\cdots<t_N=1$, offsets $o_p$ and $o_a$, readout window $G$
\State keep the prompt states fixed and draw $z_i\sim\mathcal{N}(0,\sigma^2 I)$ for every target position $i$
\For{$j=0,\ldots,N+\max(o_p,o_a)-1$}
 \State $k_s\gets\min(\max(j-o_s,0),N)$ and $t_s\gets t_{k_s}$ for $s\in\{p,a\}$
 \State predict clean states $\hat{x}_i$ for all target positions with region times $t_p$ and $t_a$ and classifier-free guidance
 \For{every target position $i$ whose region $s=s(i)$ satisfies $o_s\leq j<o_s+N$}
 \State $z_i\gets z_i+(t_{k_s+1}-t_{k_s})\,(\hat{x}_i-z_i)/\max(1-t_s,\tau)$
 \EndFor
\EndFor
\State decode the full canvas to tokens once and read the first $G$ target positions
\State \Return the answer from the answer region and the plan text for diagnostics
\end{algorithmic}
\end{algorithm}

\section{Experiments}
\label{sec:experiments}

The experiments test the claim of the paper in three steps.
Section~\ref{sec:exp-boundary} asks whether a fixed boundary makes the trace more useful and whether a fixed answer position alone explains the result.
Section~\ref{sec:exp-clocks} asks what segment clocks add to the fixed layout and in which order the plan and the answer should be denoised.
Section~\ref{sec:exp-budget} checks that the gain does not depend on a small sampling budget or on a tuned plan capacity.

\subsection{Benchmarks and comparison protocol}

Two graph reasoning benchmarks carry every comparison.
ProsQA asks whether an entity reaches one of two candidate concepts through a graph of stated implications, and every training question comes with a proof~\citep{hao2024coconut}.
ProsQA has 17,886 training, 300 validation, and 500 test questions.
We build Deep ProsQA to test longer proofs.
Every Deep ProsQA question has two chains of equal length, so the length of a chain does not reveal the answer, and proof lengths are balanced from 4 to 12 steps.
Deep ProsQA has 20,000 training, 500 validation, and 1,000 test questions, and Appendix~\ref{app:data} describes the generator.

Matched target layouts isolate the allocation change.
Direct answer predicts the answer alone, and full trace predicts the supplied proof sentences before the answer.
Free trace writes the compact trace ahead of the answer, and answer then trace reverses the order of the same two text components.
Both Plan Canvas variants use the same compact trace, canvas length, and readout window as free trace.
Table~\ref{tab:budgets} lists the canvas length and readout window of every layout.
The single-clock variant keeps the original ELF interface, and the segment-clock variant adds markers and region-dependent time conditioning.

Every layout starts from the same ELF-L initialization and frozen T5-small encoder, with about 652 million trainable backbone parameters.
Training uses thirteen epochs, a global batch size of 32, Muon optimization~\citep{liu2025muon}, and the exponential moving average of the final checkpoint.
The default sampler uses 64 synchronous Euler iterations, a classifier-free guidance scale of $2$, and a self-conditioning guidance scale of $1$.
An answer is correct when the predicted concept matches the correct candidate exactly.
Each configuration is one training run, and paired intervals describe variation over the fixed test questions of that run.
Checkpoints are not selected on test questions, and on the validation splits the two Plan Canvas variants also score above every other layout, as Table~\ref{tab:validation} in Appendix~\ref{app:stats} shows.

\begin{table}[t]
\centering
\caption{Sequence budgets in token positions, shared by training and evaluation within each target layout.
Prompt caps are 704 positions for ProsQA and 768 positions for Deep ProsQA across all layouts.
The readout window is computed from these fixed caps and never from a reference answer's content or length.}
\label{tab:budgets}
\footnotesize
\begin{tabular}{lrrrr}
\toprule
 & \multicolumn{2}{c}{ProsQA} & \multicolumn{2}{c}{Deep ProsQA} \\
Layout & Full canvas $S$ & Readout $G$ & Full canvas $S$ & Readout $G$ \\
\midrule
Direct answer & 720 & 16 & 784 & 16 \\
Full trace & 800 & 96 & 960 & 192 \\
Free trace / answer then trace & 752 & 48 & 848 & 80 \\
Plan Canvas, either clock interface & 752 & 48 & 848 & 80 \\
\bottomrule
\end{tabular}
\end{table}

\subsection{A fixed boundary on graph reasoning}
\label{sec:exp-boundary}

\begin{table}[t]
\centering
\caption{Answer accuracy and path checks on the test splits, in percent of all test questions.
A valid path starts at a fact about the entity, follows directed edges of the question graph, and ends at the predicted concept.
Correct / invalid counts correct answers with an invalid or missing path.
Free trace, answer then trace, and both Plan Canvas variants share the trace text, canvas length, and readout window, and shaded rows use the fixed plan region.}
\label{tab:main}
\footnotesize
\setlength{\tabcolsep}{3.5pt}
\renewcommand{\arraystretch}{1.12}
\begin{tabular}{@{}lrrrrrr@{}}
\toprule
 & \multicolumn{3}{c}{ProsQA} & \multicolumn{3}{c}{Deep ProsQA} \\
\cmidrule(lr){2-4}\cmidrule(l){5-7}
Target layout & Accuracy & Valid path & Correct / invalid & Accuracy & Valid path & Correct / invalid \\
\midrule
Direct answer & 87.6 & -- & -- & 57.7 & -- & -- \\
Full trace & 91.0 & 79.8 & 12.2 & 73.5 & 58.5 & \textbf{20.3} \\
Free trace & 91.4 & 75.0 & 17.0 & 73.0 & 30.8 & 42.5 \\
Answer then trace & 88.2 & 79.4 & 8.8 & 80.2 & 38.5 & 41.7 \\
\midrule
\rowcolor{planblue!7}
Plan Canvas, single clock & \textbf{96.8} & \textbf{90.6} & \textbf{6.2} & 82.1 & 55.0 & 27.1 \\
\rowcolor{planblue!7}
Plan Canvas, segment clocks & 95.2 & 89.0 & \textbf{6.2} & \textbf{87.0} & \textbf{59.1} & 27.9 \\
\bottomrule
\end{tabular}

\end{table}

Table~\ref{tab:main} gives the main comparison.
On both benchmarks the free trace improves on direct answering but not on the full trace: the free trace scores 91.4\% on ProsQA and 73.0\% on Deep ProsQA, and the full trace scores 91.0\% and 73.5\%.
Placing the same compact trace in a fixed region raises accuracy to 96.8\% and 82.1\% with the single-clock interface.
The paired 95\% intervals over the free trace are $[3.0,8.0]$ and $[6.0,12.2]$ points, with exact McNemar $p<10^{-4}$ on both benchmarks.
The two layouts share the backbone, the trace text, the canvas length, and the readout window.
What changes is the position of the answer, the supervised padding, and the target embeddings that the frozen encoder computes over the padded target.

A fixed answer position alone does not explain the gain.
The answer-then-trace control starts the answer directly after the prompt and writes the trace freely after it, so its answer position is fixed but its trace is not.
On ProsQA the control scores 88.2\%, below the free trace, and the single-clock canvas exceeds it by 8.6 points with a paired interval of $[5.6,11.6]$.
On Deep ProsQA the control comes within 1.9 points of the single-clock canvas, with an interval of $[-0.9,4.7]$, but the control reaches this accuracy without valid paths: 41.7\% of its questions are correct with an invalid or missing path, against 27.1\% for the single-clock canvas.
With segment clocks the canvas reaches 87.0\% on Deep ProsQA, 6.8 points above the control.

The fixed region also makes the emitted traces valid more often.
On Deep ProsQA the free trace gives a valid path for 30.8\% of questions, and 42.5\% of its questions are correct without one.
Plan Canvas raises valid paths to 55.0\% with the single clock and to 59.1\% with segment clocks.
The segment-clock canvas matches the path validity of the full trace, 59.1\% against 58.5\%, with a readout window of 80 positions instead of 192.
On ProsQA valid paths rise from 75.0\% to 90.6\%.

\begin{figure}[t]
\centering
\includegraphics[width=\linewidth]{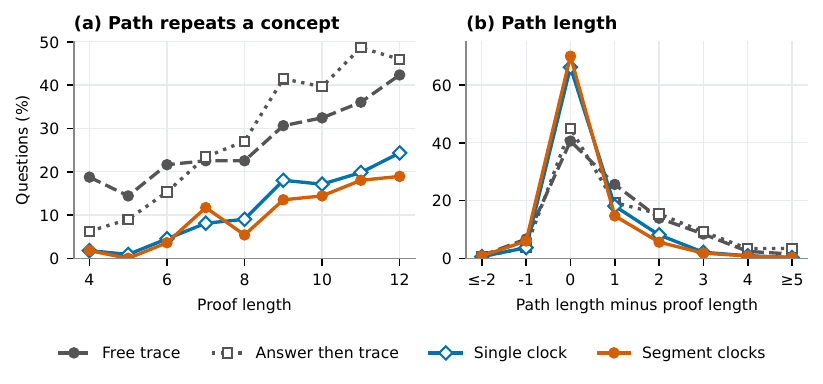}
\caption{Shape of the emitted paths on the Deep ProsQA test split, in percent of all test questions.
Panel (a) counts paths that contain the same concept twice, grouped by shortest proof length.
Panel (b) gives the distribution of the emitted path length minus the shortest proof length.}
\label{fig:shape}
\end{figure}

The failed free traces have a specific shape.
Figure~\ref{fig:shape} reads the emitted concept sequence of every Deep ProsQA output.
Almost every invalid free-trace path contains a hop that is not an edge of the question graph, and the path is usually too long: 51.8\% of free-trace paths are longer than the proof, against 29.5\% for the single clock and 23.2\% for segment clocks.
Many of the extra concepts repeat a part of the path that is already written, as in the sequence \emph{shimpus chepus shimpus chepus}.
Such repeats appear in 26.8\% of free-trace outputs and in 11.5\% and 9.7\% of the outputs of the two Plan Canvas variants.
The repeat rate of the free trace grows from 19\% at four steps to 42\% at twelve steps.
The answer-then-trace control fixes the answer position but not the end of its trace, and the control repeats concepts in 28.5\% of its outputs, as often as the free trace.
The repeats are consistent with the problem of Section~\ref{sec:method}: each trace token is placed relative to an end that is not yet decided.
The fixed plan region removes most of them.
Table~\ref{tab:example} shows one Deep ProsQA question with a proof of six steps.
The free trace follows the proof for five concepts, leaves the chain at rupus, writes scapus twice, jumps to broopus without an edge, and ends at the wrong candidate.
Both Plan Canvas variants write the six-step proof, fill the rest of the plan region with padding, and answer correctly.

The fixed region also makes the plan end at the right place.
On Deep ProsQA, the emitted path has exactly the length of the shortest proof in 40.6\% of free-trace outputs and 45.0\% of answer-then-trace outputs, against 66.2\% and 70.0\% for the two Plan Canvas variants.
On ProsQA the proofs are shorter, and the same measure rises from 74.2\% for the free trace to 89.6\% and 89.8\%.
Repeats are rare on ProsQA under every layout, at most 2.4\% of outputs, which is consistent with the smaller gain on that benchmark.
On both benchmarks the supervised padding tells the model where the plan ends, and the model writes fewer extra concepts before the answer.

\begin{table}[t]
\centering
\caption{Emitted plans for Deep ProsQA test question 450, which asks whether Eva is a chepus or a pripus among 40 stated facts and rules.
The underlined pair repeats a concept and then makes a hop that is not an edge of the question graph.}
\label{tab:example}
\footnotesize
\begin{tabular}{@{}p{0.2\linewidth}p{0.62\linewidth}p{0.1\linewidth}@{}}
\toprule
Target & Emitted plan & Answer \\
\midrule
Proof & fimpus, nulpus, molpus, slilpus, rupus, chepus & chepus \\
Free trace & fimpus nulpus molpus slilpus rupus scapus \underline{scapus broopus} pripus & pripus \\
Single clock & fimpus nulpus molpus slilpus rupus chepus, then padding & chepus \\
Segment clocks & fimpus nulpus molpus slilpus rupus chepus, then padding & chepus \\
\bottomrule
\end{tabular}

\end{table}

\begin{figure}[t]
\centering
\includegraphics[width=\linewidth]{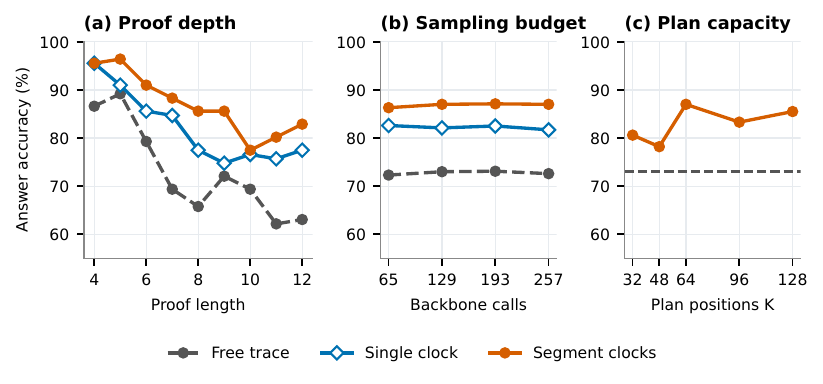}
\caption{Deep ProsQA test accuracy under three changes.
Panel (a) groups the test split by shortest proof length, with 111 or 112 questions per length.
Panel (b) evaluates the same checkpoints with 32, 64, 96, and 128 synchronous Euler iterations, counting two guided backbone calls per iteration and one decoder call.
Panel (c) trains the segment-clock canvas with different plan capacities, and the dashed line marks the free trace.}
\label{fig:diag}
\end{figure}

The gain is largest on long proofs.
Figure~\ref{fig:diag}(a) groups the Deep ProsQA test split by shortest proof length, with 111 or 112 questions per length.
On proofs of 4 to 7 steps the segment-clock canvas leads the free trace by 11.7 points, and on proofs of 8 to 12 steps it leads by 15.9 points.
The single-clock lead grows in the same way, from 8.1 to 9.9 points.
Longer proofs need longer traces, so in a free trace the answer position moves most on these questions.
The depth trend is consistent with a cost of that movement.

\subsection{Segment clocks and denoising order}
\label{sec:exp-clocks}

\begin{table}[t]
\centering
\caption{Deep ProsQA accuracy in percent as the parts of the segment-clock interface are added under tied training times.
Fixed denotes the plan region, and Marker, Prefix, and Time denote segment markers, time-prefix groups, and the position-dependent time term.}
\label{tab:ablation}
\small
\setlength{\tabcolsep}{5pt}
\begin{tabular}{@{}lccccr@{}}
\toprule
Configuration & Fixed & Marker & Prefix & Time & Deep ProsQA \\
\midrule
Free trace & -- & -- & -- & -- & 73.0 \\
Fixed plan region & $\checkmark$ & -- & -- & -- & 82.1 \\
+ segment markers & $\checkmark$ & $\checkmark$ & -- & -- & 84.0 \\
+ time-prefix groups & $\checkmark$ & $\checkmark$ & $\checkmark$ & -- & 85.9 \\
\rowcolor{planblue!8}
+ position time & $\checkmark$ & $\checkmark$ & $\checkmark$ & $\checkmark$ & 87.0 \\
\bottomrule
\end{tabular}

\end{table}

\begin{table}[t]
\centering
\caption{Accuracy in percent under different offsets with the same weights and 64 active steps per region.
Joint counts Euler iterations, Calls counts backbone evaluations including the decoder, and mixed training shares region times for half the examples.}
\label{tab:schedules}
\small
\setlength{\tabcolsep}{5pt}
\begin{tabular}{@{}lllrrrr@{}}
\toprule
Train times & Schedule & $(o_p,o_a)$ & Joint & Calls & ProsQA & Deep \\
\midrule
\rowcolor{planblue!7}
Tied & Synchronous & $(0,0)$ & 64 & 129 & 95.2 & 87.0 \\
Tied & Plan-first & $(0,64)$ & 128 & 257 & 96.0 & 87.6 \\
Tied & Answer-first & $(64,0)$ & 128 & 257 & 94.6 & 85.4 \\
\midrule
\rowcolor{planblue!7}
Mixed & Synchronous & $(0,0)$ & 64 & 129 & 95.2 & 87.9 \\
Mixed & Plan-lead & $(0,32)$ & 96 & 193 & 95.4 & 89.0 \\
Mixed & Plan-first & $(0,64)$ & 128 & 257 & 95.4 & 88.7 \\
Mixed & Answer-first & $(64,0)$ & 128 & 257 & 94.2 & 85.6 \\
\bottomrule
\end{tabular}

\end{table}

Segment clocks add to the fixed layout on Deep ProsQA.
The segment-clock canvas scores 87.0\% against 82.1\% for the single clock, a paired difference of 4.9 points with an interval of $[2.6,7.2]$ and $p=4.5\times10^{-5}$.
On ProsQA the single clock already answers 96.8\% of questions, and the segment-clock canvas scores 95.2\%.
Table~\ref{tab:ablation} adds the parts of the interface one at a time.
Segment markers, time-prefix groups, and the position-dependent time term raise Deep ProsQA from 82.1\% to 84.0\%, 85.9\%, and 87.0\%.
Each step adds 1.1 to 1.9 points and is not significant alone, with paired $p$ between 0.11 and 0.29.
Every part tells the shared Transformer more about the region of a position and the progress of that region, and together the parts give the 4.9-point gain.

Segment clocks also let one set of weights denoise the plan and the answer in different orders.
Table~\ref{tab:schedules} changes the offsets and keeps 64 active steps per region.
Under tied training, finishing the plan first scores 87.6\% on Deep ProsQA against 87.0\% for synchronous updates, a difference of 0.6 points with an interval of $[-0.7,1.9]$, and it needs twice as many joint iterations.
Starting the answer first lowers accuracy to 85.4\%, a difference of $-1.6$ points with $p=0.023$.
Mixed training repeats the pattern: plan-first scores 88.7\%, answer-first 85.6\%, and synchronous updates 87.9\%, and the answer-first drop has $p=0.003$.
The answer therefore benefits from a plan that is still being denoised, and denoising the answer before the plan costs accuracy.
A half-length plan lead under mixed training scores 89.0\%, 1.1 points above synchronous updates with $p=0.108$, at 193 instead of 129 backbone calls.
Synchronous sampling with tied training stays the default.

\subsection{Sampling budget and plan capacity}
\label{sec:exp-budget}

Two other factors could explain the gap to the free trace: the number of denoising iterations and the plan capacity.
Figure~\ref{fig:diag}(b) evaluates the same Deep ProsQA checkpoints with 65 to 257 backbone calls.
The free trace stays between 72.3\% and 73.1\%, the single-clock canvas between 81.7\% and 82.6\%, and the segment-clock canvas between 86.3\% and 87.1\%.
Four times as many calls close no part of the gap.
Figure~\ref{fig:diag}(c) trains the segment-clock canvas with 32 to 128 plan positions and sixteen answer positions.
Every capacity scores above the free trace.
The default of 64 positions rounds the longest training trace up to a multiple of eight and scores highest, at 87.0\%.
Capacities of 32 and 48 positions score 80.6\% and 78.2\%, and the 32-position region truncates 30.3\% of the training traces.
Capacities of 96 and 128 positions add padding and score 83.3\% and 85.5\%.

The fixed region does not make a sampling step more expensive.
Plan Canvas uses the canvas length and readout window of the free trace, 848 and 80 positions on Deep ProsQA, so one backbone call processes the same number of positions in both layouts.
The full trace needs a canvas of 960 positions and a readout window of 192 positions on the same benchmark, because the proof sentences are longer than the compact trace.
The segment-clock canvas is 13.5 points more accurate than the full trace and matches its path validity with 112 fewer canvas positions.
Segment clocks add 1.6 million parameters to a backbone of about 652 million parameters, and the synchronous schedule keeps the 129 backbone calls of the single-clock sampler.

\section{Related Work}

Intermediate computation can live in text~\citep{xu2025chainofdraft,zhou2024self}, in filler tokens that carry no text~\citep{pfau2024let}, in continuous states~\citep{cheng2024compressed,deng2023implicit,wu-etal-2025-parallel,xu2025softcot,wang2026plat,zhang2026soft,geiping2026scaling}, or in latent blocks.
Chain-of-thought prompting writes intermediate steps as text before the answer~\citep{wei2022chain,kojima2022large}.
Coconut feeds continuous thought states back into the model, and CODI distills reasoning into continuous states through self-distillation~\citep{hao2024coconut,shen-etal-2025-codi}.
LaDiR represents reasoning steps as compact latent blocks and refines each block with diffusion~\citep{kang2025ladir}.
Plan Canvas keeps the trace as supervised text embeddings and changes where that text lives on the canvas.

Diffusion and flow models allow every part of a sequence to be revised during generation~\citep{li2022diffusionlm,gong2023diffuseq,strudel2022self,dieleman2022cdcd,mahabadi2024tess,hu2026elf}.
Diffusion of Thought denoises reasoning traces iteratively and examines the resulting reasoning behavior~\citep{ye2024diffusionthought}.
ELF learns continuous embedding flows with a shared Transformer that performs the final token decoding~\citep{hu2026elf}.
CCDD couples continuous representations and discrete tokens in one denoising process~\citep{zhou2025ccdd}.
Diffusion Forcing trains with position-dependent noise levels~\citep{chen2024diffusionforcing}.
Flow Matching supplies the continuous-time formulation behind the region-specific updates in our implementation~\citep{lipman2023flowmatching}.
Our segment clocks apply position-dependent time to two supervised regions of one canvas, with tied training times as the default and independent times as a variant.
Spatial interfaces~\citep{kim2026any,wu2026dreamon} and commitment rules~\citep{wang2026remasking,liu2025ddpd} shape parallel language generation, and \citet{zhong2026parallelism} measure how much parallelism and which generation orders masked diffusion models use in practice.
Planned Diffusion predicts semantic chunks before denoising them in parallel, and ReFusion alternates slot selection with autoregressive infilling~\citep{israel2025planned,li2025refusion}.
\citet{kim2026interface} examine positional uncertainty through insertion-based and latent-space interfaces for masked diffusion language models.
\citet{yeom2026answer} study how answer commitment interacts with reasoning text, including interventions that delay a reserved answer position.
Semi-autoregressive and block diffusion models fix the boundaries between blocks of the output~\citep{han2023ssd,arriola2025blockdiffusion}.
In these models the trace can still end inside any block, so the answer start still depends on the trace length.
These works decide where generated text goes or when the answer is committed.
Plan Canvas fixes the boundary between a supervised intermediate trace and its answer inside one continuous flow.
The trace content and the available positions stay fixed in that measurement.

\section{Conclusion}

A free trace written into a jointly denoised canvas makes the answer start depend on a trace length that the model is still generating.
Plan Canvas removes this dependence with a fixed-capacity plan region, supervised padding, and a fixed answer position, and the fixed regions allow a separate denoising clock for the plan and for the answer.
With the trace text, backbone, and canvas length held fixed, Plan Canvas raises accuracy on ProsQA and Deep ProsQA, raises the share of deep questions answered with a valid path from 30.8\% to 59.1\%, and helps most on the longest proofs.

The emitted traces show how the layout changes generation.
Free traces often repeat part of the path and grow longer than the proof, and these repeats become more frequent as proofs get longer.
A fixed answer position alone does not remove them, because the answer-then-trace control repeats concepts as often as the free trace.
The fixed plan region does remove most of them, and it does so without a longer canvas or more backbone calls.
Once the regions are fixed, segment clocks add accuracy on deep proofs and make the order of plan and answer a choice at inference, and synchronous denoising of the two regions is the best default.
In a continuous language flow, the position of the trace on the canvas thus affects how much the same trace text helps.

\section*{AI Use Statement}

Generative AI tools were used to assist with language editing, drafting and refining portions of the manuscript, and identifying potentially relevant literature. 
All AI-assisted text and literature suggestions were reviewed and verified by the authors. 
Generative AI was not used to develop the proposed method, conduct experiments, generate experimental data, or formulate mathematical claims. 
The authors take full responsibility for the content of the paper.
\section*{Ethics Statement}

This study evaluates language models on reasoning benchmarks and does not involve experiments with human participants.

\section*{Reproducibility Statement}

Appendices document
the training settings, dataset construction, evaluation protocols,
and statistical analyses.

\bibliography{iclr2027_conference}

@article{wei2022chain,
  title={Chain-of-thought prompting elicits reasoning in large language models},
  author={Wei, Jason and Wang, Xuezhi and Schuurmans, Dale and Bosma, Maarten and Ichter, Brian and Xia, Fei and Chi, Ed H. and Le, Quoc V. and Zhou, Denny},
  journal={Advances in neural information processing systems},
  volume={35},
  pages={24824--24837},
  year={2022}
}

@article{kojima2022large,
  title={Large language models are zero-shot reasoners},
  author={Kojima, Takeshi and Gu, Shixiang Shane and Reid, Machel and Matsuo, Yutaka and Iwasawa, Yusuke},
  journal={Advances in neural information processing systems},
  volume={35},
  pages={22199--22213},
  year={2022}
}

@article{wang2023selfconsistency,
  title={Self-consistency improves chain of thought reasoning in language models},
  author={Wang, Xuezhi and Wei, Jason and Schuurmans, Dale and Le, Quoc and Chi, Ed and Narang, Sharan and Chowdhery, Aakanksha and Zhou, Denny},
  journal={arXiv preprint arXiv:2203.11171},
  year={2022}
}

@article{cheng2024compressed,
  title={Compressed chain of thought: Efficient reasoning through dense representations},
  author={Cheng, Jeffrey and Van Durme, Benjamin},
  journal={arXiv preprint arXiv:2412.13171},
  year={2024}
}

@article{xu2025chainofdraft,
  title={Chain of draft: Thinking faster by writing less},
  author={Xu, Silei and Xie, Wenhao and Zhao, Lingxiao and He, Pengcheng},
  journal={arXiv preprint arXiv:2502.18600},
  year={2025}
}

@article{deng2023implicit,
  title={Implicit chain of thought reasoning via knowledge distillation},
  author={Deng, Yuntian and Prasad, Kiran and Fernandez, Roland and Smolensky, Paul and Chaudhary, Vishrav and Shieber, Stuart},
  journal={arXiv preprint arXiv:2311.01460},
  year={2023}
}

@article{hao2024coconut,
  title={Training large language models to reason in a continuous latent space},
  author={Hao, Shibo and Sukhbaatar, Sainbayar and Su, DiJia and Li, Xian and Hu, Zhiting and Weston, Jason and Tian, Yuandong},
  journal={arXiv preprint arXiv:2412.06769},
  year={2024}
}

@inproceedings{shen-etal-2025-codi,
  title={{CODI}: Compressing chain-of-thought into continuous space via self-distillation},
  author={Shen, Zhenyi and Yan, Hanqi and Zhang, Linhai and Hu, Zhanghao and Du, Yali and He, Yulan},
  booktitle={Proceedings of the 2025 Conference on Empirical Methods in Natural Language Processing},
  pages={677--693},
  year={2025}
}

@inproceedings{wu-etal-2025-parallel,
  title={Parallel continuous chain-of-thought with {Jacobi} iteration},
  author={Wu, Haoyi and Teng, Zhihao and Tu, Kewei},
  booktitle={Proceedings of the 2025 Conference on Empirical Methods in Natural Language Processing},
  pages={914--926},
  year={2025}
}

@article{wang2026plat,
  title={Latent chain-of-thought as planning: Decoupling reasoning from verbalization},
  author={Wang, Jiecong and Peng, Hao and Liu, Chunyang},
  journal={arXiv preprint arXiv:2601.21358},
  year={2026}
}

@article{lipman2023flowmatching,
  title={Flow matching for generative modeling},
  author={Lipman, Yaron and Chen, Ricky T. Q. and Ben-Hamu, Heli and Nickel, Maximilian and Le, Matt},
  journal={arXiv preprint arXiv:2210.02747},
  year={2022}
}

@article{li2022diffusionlm,
  title={{Diffusion-LM} improves controllable text generation},
  author={Li, Xiang and Thickstun, John and Gulrajani, Ishaan and Liang, Percy and Hashimoto, Tatsunori B.},
  journal={Advances in neural information processing systems},
  volume={35},
  pages={4328--4343},
  year={2022}
}

@article{dieleman2022cdcd,
  title={Continuous diffusion for categorical data},
  author={Dieleman, Sander and Sartran, Laurent and Roshannai, Arman
          and Savinov, Nikolay and Ganin, Yaroslav and Richemond, Pierre H.
          and Doucet, Arnaud and Strudel, Robin and Dyer, Chris
          and Durkan, Conor and Hawthorne, Curtis and Leblond, R{\'e}mi
          and Grathwohl, Will and Adler, Jonas},
  journal={arXiv preprint arXiv:2211.15089},
  year={2022}
}

@article{gong2023diffuseq,
  title={{DiffuSeq}: Sequence to sequence text generation with diffusion models},
  author={Gong, Shansan and Li, Mukai and Feng, Jiangtao and Wu, Zhiyong and Kong, Lingpeng},
  journal={arXiv preprint arXiv:2210.08933},
  year={2022}
}

@article{ye2024diffusionthought,
  title={Diffusion of thought: Chain-of-thought reasoning in diffusion language models},
  author={Ye, Jiacheng and Gong, Shansan and Chen, Liheng and Zheng, Lin
          and Gao, Jiahui and Shi, Han and Wu, Chuan and Jiang, Xin
          and Li, Zhenguo and Bi, Wei and Kong, Lingpeng},
  journal={Advances in Neural Information Processing Systems},
  volume={37},
  pages={105345--105374},
  year={2024}
}

@article{chen2024diffusionforcing,
  title={Diffusion forcing: Next-token prediction meets full-sequence diffusion},
  author={Chen, Boyuan and Mart{\'\i} Mons{\'o}, Diego and Du, Yilun and Simchowitz, Max and Tedrake, Russ and Sitzmann, Vincent},
  journal={Advances in Neural Information Processing Systems},
  volume={37},
  pages={24081--24125},
  year={2024}
}

@article{hu2026elf,
  title={{ELF}: Embedded language flows},
  author={Hu, Keya and Qiu, Linlu and Lu, Yiyang and Zhao, Hanhong and Li, Tianhong and Kim, Yoon and Andreas, Jacob and He, Kaiming},
  journal={arXiv preprint arXiv:2605.10938},
  year={2026}
}

@inproceedings{kang2025ladir,
  title={{LaDiR}: Latent diffusion enhances {LLM}s for text reasoning},
  author={Kang, Haoqiang and Zhang, Yizhe and Kuang, Nikki Lijing and Majamaki, Nicklas and Jaitly, Navdeep and Ma, Yi-An and Qin, Lianhui},
  booktitle={International Conference on Learning Representations},
  volume={2026},
  pages={81705--81733},
  year={2026}
}

@inproceedings{liu2025ddpd,
  title={Think while you generate: Discrete diffusion with planned denoising},
  author={Liu, Sulin and Nam, Juno and Campbell, Andrew and St{\"a}rk, Hannes and Xu, Yilun and Jaakkola, Tommi and G{\'o}mez-Bombarelli, Rafael},
  
  booktitle={International Conference on Learning Representations},
  year={2025},
  url={https://proceedings.iclr.cc/paper_files/paper/2025/hash/cbf883a744952d4a40591271a58ab9d0-Abstract-Conference.html}
}

@article{zhou2025ccdd,
  title={Coevolutionary continuous discrete diffusion: Make your diffusion language model a latent reasoner},
  author={Zhou, Cai and Yang, Chenxiao and Hu, Yi and Wang, Chenyu and Zhang, Chubin and Zhang, Muhan and Mackey, Lester and Jaakkola, Tommi and Bates, Stephen and Zhang, Dinghuai},
  journal={arXiv preprint arXiv:2510.03206},
  year={2025}
}

@inproceedings{arriola2025blockdiffusion,
  title={Block diffusion: Interpolating between autoregressive and diffusion language models},
  author={Arriola, Marianne and Gokaslan, Aaron Kerem and Chiu, Justin T. and Yang, Zhihan and Qi, Zhixuan and Han, Jiaqi and Sahoo, Subham Sekhar and Kuleshov, Volodymyr},
  booktitle={International Conference on Learning Representations},
  volume={2025},
  pages={50726--50753},
  year={2025}
}

@article{raffel2020t5,
  title={Exploring the limits of transfer learning with a unified text-to-text transformer},
  author={Raffel, Colin and Shazeer, Noam and Roberts, Adam and Lee, Katherine and Narang, Sharan and Matena, Michael and Zhou, Yanqi and Li, Wei and Liu, Peter J.},
  journal={Journal of Machine Learning Research},
  volume={21},
  number={140},
  pages={1--67},
  year={2020}
}

@inproceedings{israel2025planned,
  title={Planned diffusion},
  author={Israel, Daniel and Jin, Tian and Cheng, Ellie and Van den Broeck, Guy and Grover, Aditya and Subramanian, Suvinay and Carbin, Michael},
  booktitle={International Conference on Learning Representations},
  volume={2026},
  pages={98298--98318},
  year={2026}
}

@inproceedings{li2025refusion,
  title={{ReFusion}: A diffusion large language model with parallel autoregressive decoding},
  author={Li, Jia-Nan and Guan, Jian and Wu, Wei and Li, Chongxuan},
  booktitle={International Conference on Learning Representations},
  volume={2026},
  pages={53846--53869},
  year={2026}
}

@article{kim2026interface,
  title={From Interface to Inference: Eliciting Any-Order Inference from Any-Order Models},
  author={Kim, Seunggeun and Kim, Jaeyeon and Lee, Taekyun and Chen, Yuyuan and Du, Yilun and Kakade, Sham and Chen, Sitan},
  journal={arXiv preprint arXiv:2607.26504},
  year={2026}
}

@article{yeom2026answer,
  title={Answer First, Reason Later: When Commitment Order Costs Accuracy in Diffusion Language Models},
  author={Yeom, Jewon and Sok, Jaewon and Park, Seonghyeon and Park, Jeongjae and Lee, Hwiyeong and Kim, Taesup},
  journal={arXiv preprint arXiv:2608.05687},
  year={2026}
}

@article{zhou2022least,
  title   = {Least-to-most prompting enables complex reasoning in large language models},
  author  = {Zhou, Denny and Sch{\"a}rli, Nathanael and Hou, Le and Wei, Jason and Scales, Nathan and Wang, Xuezhi and Schuurmans, Dale and Cui, Claire and Bousquet, Olivier and Le, Quoc and others},
  journal = {arXiv preprint arXiv:2205.10625},
  year    = {2022}
}

@inproceedings{wang2023plan,
  title     = {Plan-and-solve prompting: Improving zero-shot chain-of-thought reasoning by large language models},
  author    = {Wang, Lei and Xu, Wanyu and Lan, Yihuai and Hu, Zhiqiang and Lan, Yunshi and Lee, Roy Ka-Wei and Lim, Ee-Peng},
  booktitle = {Proceedings of the 61st Annual Meeting of the Association for Computational Linguistics (Volume 1: Long Papers)},
  pages     = {2609--2634},
  year      = {2023}
}

@inproceedings{han2023ssd,
  title     = {{SSD-LM}: Semi-autoregressive simplex-based diffusion language model for text generation and modular control},
  author    = {Han, Xiaochuang and Kumar, Sachin and Tsvetkov, Yulia},
  booktitle = {Proceedings of the 61st Annual Meeting of the Association for Computational Linguistics (Volume 1: Long Papers)},
  pages     = {11575--11596},
  year      = {2023}
}

@article{chen2022analog,
  title   = {Analog bits: Generating discrete data using diffusion models with self-conditioning},
  author  = {Chen, Ting and Zhang, Ruixiang and Hinton, Geoffrey},
  journal = {arXiv preprint arXiv:2208.04202},
  year    = {2022}
}

@article{ho2022classifier,
  title   = {Classifier-free diffusion guidance},
  author  = {Ho, Jonathan and Salimans, Tim},
  journal = {arXiv preprint arXiv:2207.12598},
  year    = {2022}
}

@article{liu2025muon,
  title   = {Muon is scalable for {LLM} training},
  author  = {Liu, Jingyuan and Su, Jianlin and Yao, Xingcheng and Jiang, Zhejun and Lai, Guokun and Du, Yulun and Qin, Yidao and Xu, Weixin and Lu, Enzhe and Yan, Junjie and others},
  journal = {arXiv preprint arXiv:2502.16982},
  year    = {2025}
}

@inproceedings{xu2025softcot,
  title     = {{SoftCoT}: Soft chain-of-thought for efficient reasoning with {LLM}s},
  author    = {Xu, Yige and Guo, Xu and Zeng, Zhiwei and Miao, Chunyan},
  booktitle = {Proceedings of the 63rd Annual Meeting of the Association for Computational Linguistics (Volume 1: Long Papers)},
  pages     = {23336--23351},
  year      = {2025}
}

@article{zhang2026soft,
  title   = {Soft thinking: Unlocking the reasoning potential of {LLM}s in continuous concept space},
  author  = {Zhang, Zhen and He, Xuehai and Yan, Weixiang and Shen, Ao and Zhao, Chenyang and Wang, Xin Eric},
  journal = {Advances in Neural Information Processing Systems},
  volume  = {38},
  pages   = {168990--169012},
  year    = {2025}
}

@article{geiping2026scaling,
  title   = {Scaling up test-time compute with latent reasoning: A recurrent depth approach},
  author  = {Geiping, Jonas and McLeish, Sean and Jain, Neel and Kirchenbauer, John and Singh, Siddharth and Bartoldson, Brian and Kailkhura, Bhavya and Bhatele, Abhinav and Goldstein, Tom},
  journal = {Advances in Neural Information Processing Systems},
  volume  = {38},
  pages   = {41340--41391},
  year    = {2025}
}

@article{strudel2022self,
  title   = {Self-conditioned embedding diffusion for text generation},
  author  = {Strudel, Robin and Tallec, Corentin and Altch{\'e}, Florent and Du, Yilun and Ganin, Yaroslav and Mensch, Arthur and Grathwohl, Will and Savinov, Nikolay and Dieleman, Sander and Sifre, Laurent and others},
  journal = {arXiv preprint arXiv:2211.04236},
  year    = {2022}
}

@inproceedings{mahabadi2024tess,
  title     = {{TESS}: Text-to-text self-conditioned simplex diffusion},
  author    = {Karimi Mahabadi, Rabeeh and Ivison, Hamish and Tae, Jaesung and Henderson, James and Beltagy, Iz and Peters, Matthew E and Cohan, Arman},
  booktitle = {Proceedings of the 18th Conference of the European Chapter of the Association for Computational Linguistics (Volume 1: Long Papers)},
  pages     = {2347--2361},
  year      = {2024}
}

@article{wang2026remasking,
  title   = {Remasking discrete diffusion models with inference-time scaling},
  author  = {Wang, Guanghan and Schiff, Yair and Sahoo, Subham and Kuleshov, Volodymyr},
  journal = {Advances in Neural Information Processing Systems},
  volume  = {38},
  pages   = {147282--147339},
  year    = {2025}
}

@inproceedings{zhong2026parallelism,
  title     = {Parallelism and generation order in masked diffusion language models: Limits today, potential tomorrow},
  author    = {Zhong, Yangyang and Gu, Yanmei and Zang, Zhengqing and Li, Xiaomeng and Ding, Yuqi and Jia, Xibei and Shen, Yuting and Lan, Zhenzhong and Zhu, Liwang and Liu, Weiping and others},
  booktitle = {Findings of the Association for Computational Linguistics: ACL 2026},
  pages     = {7178--7218},
  year      = {2026}
}

@article{wang2024chain,
  title   = {Chain-of-thought reasoning without prompting},
  author  = {Wang, Xuezhi and Zhou, Denny},
  journal = {Advances in Neural Information Processing Systems},
  volume  = {37},
  pages   = {66383--66409},
  year    = {2024}
}

@article{zhou2024self,
  title   = {{SELF-DISCOVER}: Large language models self-compose reasoning structures},
  author  = {Zhou, Pei and Pujara, Jay and Ren, Xiang and Chen, Xinyun and Cheng, Heng-Tze and Le, Quoc V. and Chi, Ed H. and Zhou, Denny and Mishra, Swaroop and Zheng, Huaixiu Steven},
  journal = {Advances in Neural Information Processing Systems},
  volume  = {37},
  pages   = {126032--126058},
  year    = {2024}
}

@article{pfau2024let,
  title   = {Let's think dot by dot: Hidden computation in {Transformer} language models},
  author  = {Pfau, Jacob and Merrill, William and Bowman, Samuel R},
  journal = {arXiv preprint arXiv:2404.15758},
  year    = {2024}
}

@inproceedings{kim2026any,
  title     = {Any-order flexible length masked diffusion},
  author    = {Kim, Jaeyeon and Lee, Cheuk-Kit and Domingo-Enrich, Carles and Du, Yilun and Kakade, Sham and Ngotiaoco, Timothy and Chen, Sitan and Albergo, Michael S.},
  booktitle = {International Conference on Learning Representations},
  volume    = {2026},
  pages     = {40151--40183},
  year      = {2026}
}

@inproceedings{wu2026dreamon,
  title     = {{DreamOn}: Diffusion language models for code infilling beyond fixed-size canvas},
  author    = {Wu, Zirui and Zheng, Lin and Xie, Zhihui and Ye, Jiacheng and Gao, Jiahui and Gong, Shansan and Feng, Yansong and Li, Zhenguo and Bi, Wei and Zhou, Guorui and others},
  booktitle = {International Conference on Learning Representations},
  volume    = {2026},
  pages     = {76626--76643},
  year      = {2026}
}
\bibliographystyle{iclr2027_conference}

\clearpage
\appendix
\section{Implementation and Training Details}
\label{app:implementation}

The experimental implementation uses the standard ELF training and generation paths with optional segment conditioning.
The backbone contains 32 Transformer blocks, hidden width 1,280, and 16 attention heads across all target layouts.
Frozen T5-small embeddings have 512 dimensions, and the learned input projection passes through a 128-dimensional bottleneck.
Embedding normalization subtracts zero and divides by 0.2 for every layout before flow training.
The segment-conditioned variants add two markers and an affine time projection, totaling 1,642,240 parameters beyond the single-clock backbone.

Clock information enters through learned prefix tokens and the position-dependent term in Equation~\ref{eq:conditioning}.
The single-clock model uses four time tokens, and the segment-conditioned model uses four tokens per region.
Both versions also use four self-conditioning guidance tokens and four model-mode tokens for the denoiser-decoder interface.
The internal plan padding uses token ID zero, while the trailing EOS padding uses token ID one.
Internal PAD positions remain part of the supervised target sequence and are removed only during final text rendering.

Fine-tuning starts from the same OpenWebText-pretrained ELF-L checkpoint at update 57,051 for every compared model.
Each configuration uses thirteen epochs and a global batch size of 32, with one update per distributed batch.
The peak learning rate is $2.5\times10^{-4}$, with 200 warmup updates, cosine decay, and zero weight decay.
Evaluation uses the final checkpoint's exponential moving average, maintained with decay 0.999 during training.
ProsQA runs finish at update 7,254, while Deep ProsQA runs finish at update 8,125.
The component and capacity runs apply the Muon update to every matrix on every device instead of splitting the matrices across devices.
These runs also reduce gradients in one bucket per update.
Both changes keep the update the same, and they keep the number of collective calls below a limit of the training cluster.
All configurations use BF16 computation with FP32 loss calculation and gradient checkpointing during training.

\subsection{Objective and self-conditioning}
\label{app:objective}

The objective assigns each training example to one branch while normalizing both branches over the same supervised positions.
For batch index $b$, let $r_b$ indicate decoder training and let $m_{bi}$ exclude prompt position $i$.
With embedding dimension $d$, the minibatch loss takes the following form in the implementation:
\begin{equation}
\mathcal{L}=\frac{\sum_{b,i}m_{bi}\left[
 r_b\,\mathrm{CE}_{bi}+(1-r_b)\frac{\|v_{\theta,bi}-v^{\star}_{\mathrm{SC},bi}\|_2^2}{d}
\right]}{\sum_{b,i}m_{bi}},
\qquad r_b\sim\mathrm{Bernoulli}(0.2).
\end{equation}
The flow branch samples times by applying the sigmoid function to a Gaussian with mean $-1.5$ and standard deviation $0.8$.
The decoder branch corrupts each position independently, using sigmoid-Gaussian coefficients with mean parameter $0.8$ and standard deviation parameter $0.8$.
Its Gaussian noise scale is $1$, compared with the flow branch's noise scale of $2$.
Decoder mode receives an endpoint time of one and a zero self-conditioning input for the final discrete reconstruction objective.

Self-conditioning retains the guidance-adjusted velocity target used by the underlying ELF implementation.
Let $v_c$ and $v_u$ denote no-gradient predictions with and without a previous clean-state estimate, respectively.
When self-conditioning is active, the target adds the following correction to the base velocity in Equation~\ref{eq:velocity}:
\begin{equation}
 v^{\star}_{\mathrm{SC}}=
 \operatorname{stopgrad}\left[v^\star+\left(1-\frac{1}{w}\right)(v_c-v_u)\right].
\end{equation}
Self-conditioning is active with probability $0.5$, and the training guidance scale $w$ ranges from $0.5$ to five.
Specifically, the implementation samples $1+w$ uniformly in log space between the corresponding endpoints 1.5 and six.
When self-conditioning is inactive, its correction and previous-estimate input are both set to zero.
Prompt dropout uses probability 0.1 across layouts to support classifier-free guidance~\citep{ho2022classifier} during conditional generation.
\subsection{Sampling budgets}

The sampled time grid contains 63 sorted interior times and the endpoints zero and one for every 64-step trajectory.
Both regions share that grid within a batch, and offsets control their respective grid indices.
Classifier-free guidance combines two backbone evaluations per joint iteration at guidance scale $2$.
The single final decoder evaluation adds one further backbone call after the denoising trajectory ends.
Thus, synchronous, half-lead, and sequential schedules require 129, 193, and 257 backbone calls, respectively, excluding the frozen encoder.

\FloatBarrier \section{Dataset Construction and Target Layouts}
\label{app:data}

All target layouts derive from the same selected proof for each question in the training split.
Direct answer uses only the answer sentence, and full trace places the supplied proof sentences before that answer.
Free trace lists the visited concept names before the answer, while answer then trace reverses these two text components.
Plan Canvas separately tokenizes the path and answer, inserts internal padding to capacity $K$, and concatenates their token sequences.
T5 decoding removes special tokens, so the internal padding is absent from the final visible path.

The plan capacity follows training token lengths and remains fixed across validation and test questions.
The longest training paths contain 25 tokens on ProsQA and 57 tokens on Deep ProsQA.
Rounding these maxima upward yields capacities of 32 and 64 plan positions, respectively.
The corresponding training proof sentences reach 87 and 179 tokens, including the final answer in each case.
Table~\ref{tab:budgets} supplies sufficient fixed readout space for every training target without using a reference answer at inference.

Deep ProsQA matches the lengths of the correct and alternative chains at every chosen proof depth.
For proof length $L$, each side has $L$ concept nodes and eight extra dead-end nodes, giving $2L+16$ concepts overall.
Each extra node receives one or two parents from its own chain or earlier extras on the same side.
Additional directed edges run from the alternative side toward the correct side while preserving the incorrect candidate's unreachability.
Entity facts include each chain's first concept and additional facts about dead-end nodes on the corresponding side.
Fact order and candidate order are shuffled, and concept names are sampled without replacement within a question.

Graph checks verify the intended reasoning depth for every generated question before it enters a dataset split.
A breadth-first search starts from all stated facts about the queried entity and measures the shortest route to the correct candidate.
The recorded depth counts the initial fact as one step, followed by the required directed implications.
Separate split generators use seeds 2,000, 2,001, and 2,002 for training, validation, and test, respectively.
Exact duplicate question strings are rejected across the three splits, and proof depths are balanced within each split.
The equal-chain variant supplies every Deep ProsQA result in this paper, and earlier generator variants served as development data.

Capacity ablations change how much trace fits before the supervised answer begins.
The Deep ProsQA sweep uses 32, 48, 64, 96, or 128 plan positions and keeps sixteen answer positions in every configuration.
If a training trace exceeds capacity, the target keeps the leading plan tokens and the complete answer.
The full sequence length grows with capacity, and every capacity configuration receives the same training examples and optimizer updates.
Sampling-budget sweeps keep the checkpoint and token budgets and vary only the number of synchronous Euler updates.

\FloatBarrier \section{Additional Graph Results and Statistical Details}
\label{app:stats}

Graph validation and test scores use the same final checkpoint and decoding settings for each fitted model.
Table~\ref{tab:validation} reports marginal Wilson intervals to describe accuracy uncertainty over the evaluated test questions.
The deeper dataset contains 500 validation questions and 1,000 test questions, while ProsQA contains 300 and 500, respectively.
These intervals characterize variation over questions and do not estimate variation across training seeds.

Paired comparisons operate on aligned correctness indicators for the complete question set of each benchmark.
We first verify unique question identifiers, reference concepts, saved predictions, aggregate counts, and final checkpoint identifiers.
For the first model and second model, let $W$ count first-only successes and $L$ count second-only successes.
The exact McNemar test compares $W$ wins against $L$ losses under equal probabilities for either discordant outcome.
Bootstrap intervals resample the paired question differences, using 20,000 replicates with analysis seed 20,260,905.
Table~\ref{tab:paired} reports all tested contrasts without multiple-comparison adjustment, and the comparisons describe the existing trained configurations.

\begin{table}[H]
\centering
\caption{Validation and test answer accuracy in percent, with P denoting ProsQA and D denoting Deep ProsQA.
Bracketed values give marginal 95\% Wilson intervals for the complete test splits under each fixed checkpoint.}
\label{tab:validation}
\footnotesize
\setlength{\tabcolsep}{3pt}
\begin{tabular}{lrrrr}
\toprule
Layout & P valid & P test [95\% CI] & D valid & D test [95\% CI] \\
\midrule
Direct answer & 88.0 & 87.6 [84.4, 90.2] & 58.2 & 57.7 [54.6, 60.7] \\
Full trace & 90.3 & 91.0 [88.2, 93.2] & 78.6 & 73.5 [70.7, 76.1] \\
Free trace & 89.3 & 91.4 [88.6, 93.5] & 71.0 & 73.0 [70.2, 75.7] \\
Answer then trace & 90.3 & 88.2 [85.1, 90.7] & 78.2 & 80.2 [77.6, 82.5] \\
Plan Canvas (single clock) & 94.0 & 96.8 [94.9, 98.0] & 82.0 & 82.1 [79.6, 84.4] \\
Plan Canvas (segment clocks) & 94.3 & 95.2 [93.0, 96.8] & 86.8 & 87.0 [84.8, 88.9] \\
\bottomrule
\end{tabular}

\end{table}

\FloatBarrier
\begingroup
\setlength{\tabcolsep}{3pt}
\setlength{\LTcapwidth}{\linewidth}
\begin{longtable}{>{\small}l>{\small}l>{\small}r>{\small}r>{\small}r>{\small}r}
\caption{\normalsize Paired test contrasts with first minus second accuracy differences, with P denoting ProsQA and D denoting Deep ProsQA.
Differences and interval endpoints use percentage points, while W/L counts discordant questions won or lost by the first model.
Canvas single denotes the single-clock canvas, segment variants specify tied or mixed training, and sync denotes synchronous generation.
The percentile bootstrap interval and exact discrete test need not cross their respective thresholds at the same contrast.
The last three rows are the component steps of Table~\ref{tab:ablation}, whose bootstrap uses 10,000 replicates with seed 0.}\label{tab:paired}\\
\toprule
Set & First / second & $\Delta$ (pp) & Paired 95\% CI & W/L & $p$ \\
\midrule
\endfirsthead

\multicolumn{6}{c}{\tablename~\thetable{} (continued)} \\
\toprule
Set & First / second & $\Delta$ (pp) & Paired 95\% CI & W/L & $p$ \\
\midrule
\endhead

\midrule
\multicolumn{6}{r}{\small Continued on next page} \\
\endfoot

\bottomrule
\endlastfoot

P & Canvas single / Free trace & +5.4 & [+3.0, +8.0] & 34/7 & $2.53\times10^{-5}$ \\
P & Canvas single / Full trace & +5.8 & [+3.2, +8.4] & 37/8 & $1.54\times10^{-5}$ \\
P & Canvas single / Answer then trace & +8.6 & [+5.6, +11.6] & 52/9 & $1.80\times10^{-8}$ \\
P & Segment tied, sync / Canvas single & -1.6 & [-3.0, -0.2] & 3/11 & 0.057 \\
P & Segment mixed, sync / Segment tied, sync & +0.0 & [-1.4, +1.4] & 7/7 & 1.000 \\
P & Tied, Plan-first / Segment tied, sync & +0.8 & [+0.0, +1.8] & 5/1 & 0.219 \\
P & Tied, Answer-first / Segment tied, sync & -0.6 & [-1.6, +0.4] & 2/5 & 0.453 \\
P & Mixed, Plan-lead / Segment mixed, sync & +0.2 & [-1.0, +1.4] & 5/4 & 1.000 \\
P & Mixed, Plan-first / Segment mixed, sync & +0.2 & [-1.0, +1.4] & 5/4 & 1.000 \\
P & Mixed, Answer-first / Segment mixed, sync & -1.0 & [-2.4, +0.2] & 3/8 & 0.227 \\

\addlinespace[3pt]

D & Canvas single / Free trace & +9.1 & [+6.0, +12.2] & 173/82 & $1.23\times10^{-8}$ \\
D & Canvas single / Full trace & +8.6 & [+5.1, +12.0] & 200/114 & $1.40\times10^{-6}$ \\
D & Canvas single / Answer then trace & +1.9 & [-0.9, +4.7] & 110/91 & 0.204 \\
D & Segment tied, sync / Canvas single & +4.9 & [+2.6, +7.2] & 95/46 & $4.49\times10^{-5}$ \\
D & Segment mixed, sync / Segment tied, sync & +0.9 & [-1.0, +2.8] & 49/40 & 0.397 \\
D & Tied, Plan-first / Segment tied, sync & +0.6 & [-0.7, +1.9] & 26/20 & 0.461 \\
D & Tied, Answer-first / Segment tied, sync & -1.6 & [-2.9, -0.3] & 14/30 & 0.023 \\
D & Mixed, Plan-lead / Segment mixed, sync & +1.1 & [-0.1, +2.3] & 25/14 & 0.108 \\
D & Mixed, Plan-first / Segment mixed, sync & +0.8 & [-0.5, +2.1] & 27/19 & 0.302 \\
D & Mixed, Answer-first / Segment mixed, sync & -2.3 & [-3.8, -0.9] & 16/39 & 0.003 \\

\addlinespace[3pt]

D & Markers, sync / Canvas single & +1.9 & [-0.3, +4.2] & 77/58 & 0.121 \\
D & Time-prefix groups, sync / Markers, sync & +1.9 & [-0.3, +4.1] & 73/54 & 0.110 \\
D & Segment tied, sync / Time-prefix groups, sync & +1.1 & [-0.7, +2.9] & 50/39 & 0.289 \\

\end{longtable}

\endgroup

Reconstructed path checks distinguish graph consistency from whether the answer matches the correct candidate.
Table~\ref{tab:main} reports the proportions of all test questions that have valid paths and correct answers accompanied by invalid or missing paths.
For concept sequences, reconstruction reads the concept names surrounding the answer sentence according to the target layout.
For full proof text, reconstruction extracts the entity's initial concept and the implication endpoints in their generated order.
The resulting concept sequence must satisfy the start, directed-edge, and answer-endpoint checks used throughout the paper.
This procedure measures consistency of recovered paths, and a valid path can still end at an incorrect or irrelevant concept.

\end{document}